\documentclass{article}

\usepackage{arxiv}

\usepackage[utf8]{inputenc} 
\usepackage[T1]{fontenc}    
\usepackage{hyperref}       
\usepackage{amsfonts}       
\usepackage{nicefrac}       
\usepackage{microtype}      
\usepackage{lipsum}
\usepackage{graphicx}
\graphicspath{ {./images/} }

\usepackage[numbers]{natbib}
\usepackage{amsmath}
\usepackage{amsthm}
\usepackage{comment}
\usepackage[table]{xcolor}
\usepackage{tabularx}

\usepackage{url}

\usepackage{enumitem}

\usepackage{booktabs} 
\usepackage{breqn}
\usepackage{mathtools}
\usepackage[linesnumbered,ruled]{algorithm2e}

\usepackage{amssymb}
\usepackage{textcomp}
\usepackage{multirow}
\usepackage{wrapfig}

\usepackage{graphicx}

\usepackage[normalem]{ulem}
\usepackage[T1]{fontenc}
\usepackage{color,soul}

\usepackage{longtable}
\usepackage{subcaption}

\usepackage{adjustbox}

\usepackage{geometry}

\newcommand\citeA[1]{%
  \citeauthor{#1}~(\citeyear{#1})}

\title{A Survey on Physarum Polycephalum Intelligent Foraging Behaviour and Bio-Inspired Applications}

\author{
 Abubakr Awad \\
  School of Computer Science\\
  University of Nottingham\\
  Nottingham, NG8 1BB\\
  United Kingdom\\
  \texttt{abubakr.awad@nottingham.ac.uk} \\
   \And
 Wei Pang \\
  School of Mathematical and Computer Sciences\\
  Heriot-Watt University\\
  Edinburgh, EH14 4AS\\
  United Kingdom\\
  \texttt{w.pang@hw.ac.uk} \\
  \And
 David Lusseau \\
  National Institute of Aquatic Resources\\
  Technical University of Denmark\\
  2800 Kgs. Lyngby \\
  Denmark \\
  \texttt{davlu@aqua.dtu.dk} \\
  \And
 George M. Coghill \\
  School of Natural and Computing Sciences\\
  University of Aberdeen\\
  Aberdeen, AB24 3UE\\
  United Kingdom\\
  \texttt{g.coghill@abdn.ac.uk } \\
}

\begin{document}
\maketitle
\begin{abstract}
In recent years, research on Physarum polycephalum has become more popular after Nakagaki et al. (2000) performed their famous experiment showing that Physarum was able to find the shortest route through a maze. Subsequent researches have confirmed the ability of Physarum-inspired algorithms to solve a wide range of NP-hard problems. In contrast to previous reviews that either focus on biological aspects or bio-inspired applications, here we present a comprehensive review that highlights recent Physarum polycephalum biological aspects, mathematical models, and Physarum bio-inspired algorithms and their applications. The novelty of this review stems from our exploration of Physarum intelligent behaviour in competition settings. Further, we have presented our new model to simulate Physarum in competition, where multiple Physarum interact with each other and with their environments. The bio-inspired Physarum in competition algorithms proved to have great potentials for future research.
\end{abstract}

\keywords{Slime Mould \and Physarum Polycephalum \and Bio-inspired Algorithms \and Competition Modelling}

\section{Introduction\label{sec:introduction}}
Bio-inspired computing focuses on extracting computational models for problem solving from in-depth understanding of behaviour and mechanisms of biological systems. In recent years, cellular computational models based on the structure and the processes of living cells, such as bacterial colonies \cite{RefWorks:304} and viral models \cite{RefWorks:364} have become an important line of research in bio-inspired computing. Physarum-computing, as an example of cellular computing model, has attracted the attention of many researchers \cite{RefWorks:67}. Physarum polycephalum (Physarum for short) is an example of plasmodial slime moulds that are classified as a fungus "Myxomycetes" \cite{RefWorks:49}. In recent years, research on Physarum-inspired computing has become more popular since \citeA{RefWorks:34} performed their well-known experiments showing that Physarum was able to find the shortest route through a maze \cite{RefWorks:34}. Recent research has confirmed the ability of Physarum-inspired algorithms to solve a wide range of problems \cite{RefWorks:213,RefWorks:214}.

Physarum can be modelled as a reaction-diffusion system (cytoplasmic liquid) encapsulated in an elastic growing membrane of actin–myosin cytoskeleton \cite{RefWorks:73}. In the early stages of growth (i.e., the exploration phase), the Physarum foraging behaviour results in the generation of a branching pattern. In the second phase (i.e., the exploitation phase), it spans the sources of nutrients with a dynamic proximity graph and forms a pattern similar to Voronoi diagram \cite{RefWorks:345}. This characteristic of continuous change in Physarum protoplasmic flux with the change of environment allows Physarum-inspired algorithms to have great potentials in dealing with graph-optimisation problems \cite{RefWorks:213}.

Computer scientists are investigating the potential of Physarum-inspired techniques for solving many NP-hard problems \cite{RefWorks:33}. 
Physarum is capable of decision-making and information processing that can lead to the emergence of complex social behaviour \cite{RefWorks:55,RefWorks:57,RefWorks:124}. It compares the relative qualities of multiple options and combines the information on reward in order to make correct and adaptive decisions. Physarum is also capable of memorising and anticipating repeated events, and displays both short and long term habituation, as a simple form of learning \cite{RefWorks:211,RefWorks:344}.

In deed, Physarum can be considered as one of the biological example of unconventional computation capable of creating a programmable Physarum machine \cite{RefWorks:33}. It has been studied in the project "Physarum chip: growing computers from slime mould" \cite{RefWorks:326} that ran between 2013 and 2016. The Physarum chip is expected to solve a wide range of computation tasks, including graph optimisation, logic and arithmetical computing \cite{RefWorks:341}. The EU-funded project “Physarum Sensor: Biosensor for Citizen Scientists” is an extension of the PhyChip project \cite{RefWorks:358}. This project showed that Physarum is an ideal biological substrate that could be used as biosensors that convert a biological response into an electrical signal. These low-cost biosensors can be used for various applications, including environmental monitoring and health \cite{RefWorks:355}.

Several reviews on Physarum have been published, however, they either focus on biological aspects \cite{RefWorks:49,RefWorks:156} or mathematical models and bio-inspired applications \cite{RefWorks:213,RefWorks:214,RefWorks:33,RefWorks:387}. The novelty of this review stems from a comprehensive survey that summarises the latest published literature on Physarum covering biological behaviour, reflection on modelling, and computing aspects. Further, we have covered Physarum-inspired applications from three aspects. In contrast to other Physarum review papers that focus mainly on Physarum-inspired algorithms to solve graph optimisation problems, we have discussed a second aspect of application which is taking advantage of Physarum characteristics, such as morphological diversity \cite{RefWorks:76} and positive feedback loop \cite{RefWorks:260}, that will lead to the development of hybrid algorithms that optimise evolutionary algorithms to improve its efficiency and robustness \cite{RefWorks:333,RefWorks:98}. In the third aspect of Physarum applications, we demonstrated Physarum as a method of biological computing that has been extensively studied in "Physarum chip: growing computers from slime mould" \cite{RefWorks:326}, and the PhySense project "Physarum Sensor: Biosensor for Citizen Scientists" \cite{RefWorks:355}.

In this review we have presented our novel model that simulate Physarum in competition settings. To the best of our knowledge, we are the first to explore Physarum intelligent behaviour in competition settings, unlike the other models based on a single Physarum. Multiple Physarum with autonomous behaviours react to each other and with their environment, this has allowed the efficient exploration of the whole system evolving to an optimal global network and each Physarum to move to a better position. Further, it has allowed us to deal with the increasingly proposed networks scenarios with multiple sources and multiple sinks. The bio-inspired Physarum competition algorithms proved to have great potentials in dealing with graph-optimisation problems in a dynamic environment as in Mobile Wireless Sensor Networks\cite{RefWorks:368}, and Discrete Multi-Objective Optimisation problems \cite{RefWorks:307}.

We will start by giving a short overview of bio-inspired computing (Section \ref{sec:LiteratureReview/Bioinspired_Computing}). For deep understanding of Physarum biological foraging behaviour we will review the biological aspects of Physarum including its intelligent foraging behaviour, collective swarm behaviour, and competitive behaviour (Sections \ref{sec:LiteratureReview/Physarum_Biology}, \ref{sec:LiteratureReview/Physarum_Intelligent_Behaviour}, \ref{sec:LiteratureReview/Physarum_Swarm_Intelligence}, and \ref{sec:LiteratureReview/Physarum_Competitive_Foraging_Behaviour}). Then we will present some of the most well-known real biological experiments and mathematical models (Section \ref{sec:LiteratureReview/Physarum_Biological_Experiments} and Section \ref{sec:LiteratureReview/Mathematical_Models}). Furthermore, we will present some of the real-world applications that have been solved by Physarum-inspired algorithms (Section \ref{sec:LiteratureReview/Physarum_Applications}). Finally, a conclusion is given in Section \ref{sec:LiteratureReview/Conclusion}.

\section{Bio-inspired Computing \label{sec:LiteratureReview/Bioinspired_Computing}}
The inspiration from biology and nature has always been one of the most important and exhaustless sources for researchers and engineers to develop novel algorithms and innovative techniques during the past decades. Earlier works on bio-inspired computing focus on extracting the computational models from complex high-level biological systems of cognition and understanding. Under this umbrella of computational intelligence, there are many paradigms such as artificial neural networks, genetic algorithm, and artificial immune system. These models are based on imitating the behaviour of central nervous system, chromosomal reproduction, and immunity against infection, respectively \cite{RefWorks:182}. In recent years, simple cellular computational models based on the structure and the processes of living cells became an essential branch of bio-inspired computing, such as bacterial colonies \cite{RefWorks:304} and viral models \cite{RefWorks:364}. Physarum is an example of a cellular computing model attracting researchers' attention \cite{RefWorks:67}.

Two typical categories of bio-inspired algorithms are evolutionary algorithms and swarm intelligence algorithms, which are inspired by the natural evolution and collective behaviour in swarms of animals, respectively \cite{RefWorks:182}. However, there are several existing limitations of these optimisation methodologies \cite{RefWorks:359}. Evolutionary algorithms use iterative progresses in a population in response to environmental pressure that causes natural selection, and this causes an increase in the fitness of the population. Genetic Algorithm (GA) \cite{RefWorks:360} is an example of evolutionary algorithms. Swarm intelligence is one of the most exciting topics dealing with the collective behaviour of decentralised and self-organised biological systems. It consists of a population of simple agents which can communicate locally with each other and their local environment. These interactions can lead to the emergence of hugely complicated global behaviour \cite{RefWorks:182}. A variety of swarm intelligence algorithms for optimisation problems, such as particle swarm optimisation \cite{RefWorks:183}, ant colony optimisation \cite{RefWorks:184}, and Artificial Bee Colony (ABC) \cite{RefWorks:185}, have been developed with increasingly wide applications in the real world.

\section{Slime Mould (Physarum Polycephalum) Biology and Foraging Behaviour \label{sec:LiteratureReview/Physarum_Biology}}
\begin{figure*}
\includegraphics[width=\textwidth]{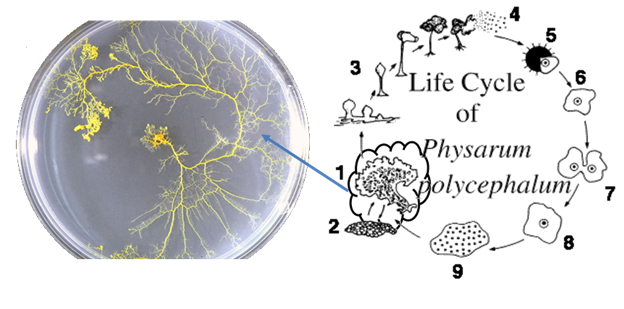}
\caption{The Physarum tube networks and life cycle}
\label{fig:Literature_Review/Physarum_Life_Cycle}
\end{figure*}

Slime mould was classified as a fungus, a class of Myxomecetes, but now it is considered to be part of kingdom Protista. There are two main types of slime moulds: the cellular slime moulds, and the plasmodial  slime moulds. The cellular slime moulds are formed  of multiple cells, whereas the plasmodial slime moulds are formed of a large multi-nucleated single cell with thousands of nuclei without any membrane between them \cite{RefWorks:156,RefWorks:49}.
Physarum polycephalum (Physarum for simplicity) is an example of plasmodial slime moulds; it consists of a single cell amoeba-like organism and has a simple structure which can be easily modelled (compared to others like ants or bees). Physarum strains are not related to fungi and form a genuine branch in the evolution tree of life, other than fungi. More than 800 slime mould species exist worldwide \cite{RefWorks:326}. This organism has a sophisticated life cycle (Figure \ref{fig:Literature_Review/Physarum_Life_Cycle}) \cite{RefWorks:49}.
The primitive intelligence of Physarum is mostly demonstrated during its vegetative stage when it turns into plasmodium. In this stage, it forms a yellowish vascular network which expands up to tens of centimetres in search of food to connect the food source (e.g., oat flakes) with the Physarum body \cite{RefWorks:50}. Physarum can be considered as a parametric bio-blob that presents itself as a geometrically smart adaptive graph structure \cite{RefWorks:345}. It is formed of a mycelial tubular network through which the chemical and physical signals, the nutrients, and the body mass are transported throughout the organism. The tubes of plasmodium Physarum are made of a gel-like outer membrane of actin–myosin cytoskeleton that generates periodic contractions of the tube walls. Inside this membrane, the cytoplasmic liquid is pumped back and forth in a rhythmically oscillating manner. The contraction amplitude and the frequency generally increase or decrease when encountering an attractant or repellent, respectively \cite{RefWorks:52,RefWorks:365}.

The Physarum senses gradients of chemoattractants and repellents and forms a yellowish vascular network in search of nutrition \cite{RefWorks:52,RefWorks:53}. It responds to stimulation by changing patterns of electrical potential oscillations, and it is made of hundreds to thousands of biochemical oscillators \cite{RefWorks:52}. A stimulus triggers the release of a signalling molecule cyclic adenosine monophosphate (cAMP) \cite{RefWorks:366} that starts cytoplasmic streaming. This stimulus gives rise to propagating waves resulting in increased cytoplasmic streaming (shuttling) through that vein \cite{RefWorks:52,RefWorks:365}. This generates a positive feedback loop; the higher the rate of cytoplasmic streaming is, the thicker the vein becomes \cite{RefWorks:367}.

The Physarum foraging behaviour consists of two simultaneous self-organised processes: expansion (exploration) and contraction (exploitation).
Physarum structure reveals two distinct geometric patterns: (a) Physarum develops thin branches, searching in their environment for food, (b) the bulging droplet-like blobs enlargement at the tips of the branches \cite{RefWorks:345}. In the early stages (exploration phase) the organism grows, and the branches with the bulging blobs at their tips become longer through the foraging process, and they divide into further branches and link up like veins. In the second phase (exploitation phase) the tubes that transport the nutrients will grow bigger while the tubes which do not transport enough nutrients will vanish and disappear \cite{RefWorks:66,RefWorks:345}.

\section{Physarum Intelligent Behaviour \label{sec:LiteratureReview/Physarum_Intelligent_Behaviour}}
Physarum may not have a brain, but computer scientists are investigating its potential as novel, unconventional computers \cite{RefWorks:33}. Physarum is capable of making complex foraging decisions based on trade-offs between risks, hunger level and food patch quality \cite{RefWorks:55,RefWorks:56,RefWorks:57,RefWorks:58,RefWorks:310}. The primitive intelligence of Physarum Polycephalum (slime mould) is mostly demonstrated during its plasmodium stage (a large multi-nucleated single cell). The underlying mechanisms of Physarum intelligence and cognition are based on the way with which the organism perceives the environment, integrates this information and makes decisions \cite{RefWorks:124}. This has motivated many researchers to take inspiration from their biological phenomena to come up with a novel, biologically inspired models for unconventional computational methods capable of solving many NP-hard problems \cite{RefWorks:33}. 
In what follows we will summarise the Physarum intelligent behaviours.

\subsection*{Finding Shortest Path}
This intelligent behaviour was first observed by \citeA{RefWorks:34} \cite{RefWorks:34}. Physarum was able to find the shortest path between two selected points (source node, and sink node) in a maze-solving problem. Other examples of the shortest path approach may include the towers of Hanoi problem \cite{RefWorks:45}.

\subsection*{Building High-Quality Networks}
Physarum's network design ability has attracted the attention of many researchers as it showed excellent ability in network construction without central consciousness during foraging process \cite{RefWorks:35,RefWorks:211,RefWorks:124}. In the early stages (exploration phase) the organism's branches grow and the bulging blobs divide into further branches and link up like veins. In the second phase (exploitation phase) the organism eventually spans the sources of nutrients with a dynamic proximity graph (Voronoi pattern), where the links (edges) connect the corresponding nodes (vertices). This network architecture is highly dynamic with flexible rearrangement of its junctions, and once the Physarum moves, the location, size of the vertices and the edges changes, disappear, or new links and vertices (nodes) develop \cite{RefWorks:345}.

One of the most well-known real experiments that showed the intelligence of Physarum for network design was the reconstruction of Tokyo railway network designed by Physarum \cite{RefWorks:67}. Some other real-world transportation networks have also been approximated by Physarum since then, such as Mexican highway \cite{RefWorks:313}, Iberian highway \cite{RefWorks:312}, Route 20 in USA \cite{RefWorks:51} and Autobahn 7 in Germany \cite{RefWorks:51}.

\subsection*{Adapting to Changing Environments}
Many biological experiments have shown that Physarum networks disassemble and reassemble within a period of a few hours in response to the change of external conditions (e.g., chemotaxis, phototaxis and thermotaxis) \cite{RefWorks:314}. Moreover, \citeA{RefWorks:315} has shown that Plasmodium-based computing devices can be precisely controlled and shaped by illumination \cite{RefWorks:315}. \citeA{RefWorks:305} have demonstrated how a growth parameter in the model can be used to transit between Convex and Concave Hulls \cite{RefWorks:305}. These results demonstrated how Physarum can approximate the external and internal shape of a set of points using chemo-attractant stimuli and masking by light illumination (repellent).

\section{Physarum Collective Behaviour and Swarm Intelligence \label{sec:LiteratureReview/Physarum_Swarm_Intelligence}}


Physarum exhibits swarm intelligence and social behaviour as social insects and animals. It shares with these insects and animals many common features of collective behaviour, such as synchronisation, communication, positive feedback, distributed intelligence, and spatial memory \cite{RefWorks:156}. Physarum's collective behaviour is the result of communication and interactions among its individual units. Being a single-cell organism, Physarum individual units do not have a ’choice’ to behave selfishly; rather, they communicate together via chemical transmitter namely cyclic adenosine monophosphate (cAMP) signals/oscillators which coordinate and synchronise Physarum's slug behaviour. A stimulus triggers the release of a signalling molecule cAMP that results in changing patterns of electrical potential oscillations which starts cytoplasmic streaming. This is distinct from other social animals such as bees or ants which use other types of communication (e.g., pheromone for ants) \cite{RefWorks:156}.

The following points will summarise the Physarum collective behaviour and swarm intelligence.

\subsection*{Synchronisation and Communication}
The plasmodium (Physarum) shows synchronous oscillation of cytoplasm throughout all its parts that behave cooperatively for exploring the space, searching for nutrients, and optimising the network of streaming protoplasm. Each tiny oscillator is a segment of a tubule network, which is actively expanding and contracting as a form of distributed, collective behaviour that allows Physarum to make complex decisions when exploring its environment. This response causes the cytoplasm to flow in the direction of the attractant and away from repellent \cite{RefWorks:53}.

\subsection*{Feedback Mechanism}
Physarum protoplasm migrates towards the area of the highest cAMP concentration and at the same time starts secreting cAMP. This behaviour creates a positive feedback loop, which will cause protoplasmic tubes with high cAMP levels to grow bigger and those with low cAMP levels to disappear gradually due to lack of flow \cite{RefWorks:66}. The tubes that are more suitable for transporting the nutrients will grow bigger and will be of less resistance. On the other hand, the tubes which do not transport enough nutrients will eventually vanish and disappear. This feedback mechanism makes Physarum intelligent enough to maximise the number of nutrient sources and minimise transportation costs \cite{RefWorks:66,RefWorks:345}. However, such positive feedback in Physarum is weaker than the ant colonies in the same maze problem, this will allow Physarum to discover and utilise new solutions and prevent the convergence on a single best solution \cite{RefWorks:60}.

\subsection*{Distributed Intelligence}
Physarum may not have a central information processing unit like a brain, but rather a collection of similar parts of protoplasm. Physarum has recently emerged as a model system for studying information processing and problem-solving in non-neuronal organisms \cite{RefWorks:356}. Physarum is a system describing the characteristics of a liquid geometry computer in conversation with its environment to survive \cite{RefWorks:345}. This type of intelligence is now considered as a part of the theme "Liquid brains: How distributed cognitive architectures process information" \cite{RefWorks:344}. Thus, Physarum is an excellent candidate for research on autonomous distributed network optimisation \cite{RefWorks:318}.

\subsection*{Memorising and Learning}
Both learning and memory are essential features for animals to survive, and information on past experiences is used for optimal decision-making in a dynamic environment. Physarum is capable of memorising and anticipating repeated events. This intelligent behaviour was first revealed by \citeA{RefWorks:58} \cite{RefWorks:58}. Moreover, \citeA{RefWorks:320} used an associate learning experiment to test this ability further \cite{RefWorks:320}. Physarum secretes a trail of slime following movement, which acts as an extra-cellular spatial memory. This increases foraging efficiency of Physarum by avoiding previously explored areas \cite{RefWorks:93,RefWorks:59}.
Physarum displays both short and long-term habituation as a simple form of learning. The information acquired during the habituation, even to chemical repellents, is via constrained absorption of these chemicals to be used as a "circulating memory" \cite{RefWorks:344}.

\section{Physarum Competitive Foraging Behaviour \label{sec:LiteratureReview/Physarum_Competitive_Foraging_Behaviour}}

\subsection*{Competition}
Competition is generally considered as negative effects caused by the presence of competitors, usually leading to the reduction of available resources. However, competition can also yield lower overall costs, better quality, more choices and varieties, more innovation, greater efficiency, and productivity \cite{RefWorks:342}.
Competition can be classified into exploitation competition and exclusion competition based on the interactions of the competitors \cite{RefWorks:371}. Exploitation competition happens when a resource that is in short supply is reduced by other competitors. This will negatively affect another competitors using the same resource. Only the more powerful competitors can obtain this limited opportunity. Exclusion competition regulates population density by slowing down the population increase if the population density is high and vice versa. Competition is very important in driving natural selection as a superior competitor can eliminate inferior ones from the area, resulting in competitive exclusion \cite{RefWorks:86}.

\subsection*{Physarum Foraging Behaviour in Competition Settings}
There is increasing evidence that a simple organism like Physarum has complex social behaviours including cooperation and competition \cite{RefWorks:56,RefWorks:346,RefWorks:316,RefWorks:193}. Physarum is capable of making complex foraging decisions based on trade-offs between risks, hunger level and food patch quality \cite{RefWorks:56}. The skills of individual competitors are effective methods for inspiration to develop intelligent systems and to provide solutions for decision-making problems. Competitions between multiple Physarum is based on Physarum power (genotype), mass, and the availability of nearby food resources \cite{RefWorks:316}. Physarum always initiates foraging behaviour quicker in the presence of competitors \cite{RefWorks:308}.

\begin{figure*}
\begin{center}
\includegraphics[scale=2.3]{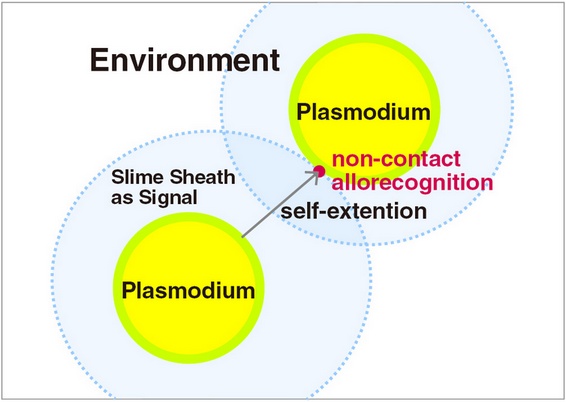}
\caption{Self-extension model with non-contact allorecognition. Self-extension occurs using the slime sheath as a signal transmitted to the environment, which facilitates non-contact allorecognition \cite{RefWorks:346}.}
\label{fig:Model/Allorecognition_self_extension_model}
\end{center}
\end{figure*}

A recent study by \citeA{RefWorks:346} \cite{RefWorks:346} has provided an answer to a crucial question: can Physarum identify allogeneic individuals? The answer is yes, allorecognition implicitly promotes the Physarum's ability of to distinguish its own tissues from those of another, when encountering different individuals.
In early research, people adopted the hypothesis that Plasmodium allorecognition was based on the premise of contact, and the slime sheath is just regarded as a simple repellent \cite{RefWorks:59,RefWorks:38}. However, the recent study by \citeA{RefWorks:346} \cite{RefWorks:346} has indicated that the slime sheath is a substance that disperses allorecognition information about itself into the environment. This view led to a new self-extension model (Figure \ref{fig:Model/Allorecognition_self_extension_model}), in which the mechanism of non-contact allorecognition using a slime sheath expands the plasmodium opportunities for decision-making, which frequently enables early and safe avoidance rather than fusion \cite{RefWorks:346}.

\section{Physarum Real Biological Experiments \label{sec:LiteratureReview/Physarum_Biological_Experiments}}
Many experiments have been made to reveal Physarum intelligence. From a computer science point of view, the objective of creating such experiments is to build a mathematical model inspired by real biological experiments to solve real-world optimisation problems.
We have summarised some of these biological experiments in Table \ref{tab:Literature_Review/real_biological_experiments}.

\begin{table}[htbp]
  \centering
  \caption{Biological Experiments, where \# PH, \# FS is the number of Physarum strains and food resources in the experiment, respectively.}
  \label{tab:Literature_Review/real_biological_experiments}
    \begin{tabular}{|l|p{6.07em}|p{8.215em}|l|p{2.07em}|l|p{6.57em}|l|p{5.57em}|}
    \toprule
    \rowcolor[rgb]{ .31,  .506,  .741}       & \textbf{Author } & \textbf{Aim } & \multicolumn{1}{p{2.355em}|}{\textbf{\# PH }} & \textbf{\# FS} & \textbf{Environment } & \textbf{Measuring Instrument } \\
    
    \rowcolor[rgb]{ .816,  .847,  .91} 1     & \citeA{RefWorks:34} & Physarum solving maze problem & 1     & 1 & Petri dish  & Camera  \\
    
    \midrule
    
    \rowcolor[rgb]{ .914,  .929,  .957} 2     & \citeA{RefWorks:67} & Physarum solving minimum spanning tree  & 1     & N     & Petri dish  & Camera  \\
   
    \midrule
    
    \rowcolor[rgb]{ .816,  .847,  .91} 3     & \citeA{RefWorks:68} & Physarum movement based on the statistical results  & 1     & 0 & CA like dish  & Camera  \\
    
    \midrule
    
    \rowcolor[rgb]{ .914,  .929,  .957} 4     & \citeA{RefWorks:122} & Physarum changes patterns of its electrical activity when exposed to attractants and repellents & 1     & 1 & Petri dish  & Electric Potential  \\
   
    \midrule
    
    \rowcolor[rgb]{ .816,  .847,  .91} 5     & \citeA{RefWorks:124} & How Physarum solves two bandit problem  & 1     & N     & Petri dish  & Camera  \\
    
    \midrule
    
    \rowcolor[rgb]{ .914,  .929,  .957} 6     & \citeA{RefWorks:308} & How Physarum tune its foraging decision when faced with competition & 2     & 1 & Petri dish  & Camera  \\
    
    \midrule
    
    \rowcolor[rgb]{ .816,  .847,  .91} 7     & \citeA{RefWorks:316} & How Physarum power (type) and mass affects foraging behaviour in competition settings & 2     & N & Petri dish  & Camera  \\
    
    \midrule
    
    \rowcolor[rgb]{ .914,  .929,  .957} 8     & \citeA{RefWorks:346} & Physarum's ability to distinguish its own tissues from those of another (Allorecognition) & 2     & 2     & Petri dish  & Camera  \\
    \bottomrule
    \end{tabular}%
\end{table}%

\subsection*{Physarum solving maze experiment}
\citeA{RefWorks:34} designed a biological experiment where a Physarum was capable of solving a maze \cite{RefWorks:34}. The goal of the experiment was to demonstrate the intelligent behaviour of a single Physarum capable of finding the shortest path between two points. In this experiment, there was only one Physarum and one food resource (i.e., solving the shortest path problem).

\subsection*{Physarum network construction experiment}
\citeA{RefWorks:67} designed a biological experiment to simulate the Physarum network formation for the Tokyo railway network and other cities \cite{RefWorks:67}. The goal of the experiment was to demonstrate the intelligent behaviour of a single Physarum (as a representation of Tokyo) capable of finding the minimum spanning tree that covers all points of multiple food resources (as a representation to other Japanese cities).

\subsection*{Physarum shape representation experiment}
The behaviour of the plasmodium is mediated by environmental stimuli. \citeA{RefWorks:50} demonstrated how a growth parameter in the model can be used to achieve transition between convex and concave hulls \cite{RefWorks:50}. These results suggested novel mechanisms of morphological computation mediated by environmental stimuli and demonstrated how Physarum polycephalum can approximate the external and internal shape of a set of points using chemo-attractant stimuli and masking by light illumination.

\subsection*{Physarum living cellular automata experiment}
The majority of these experiments have focused on Physarum behaviour in an open space (Petri dish). However, investigating the Physarum behaviour in a closed space will help us to understand how the organism makes its decision in a stepwise transition. To accomplish this goal, \citeA{RefWorks:68} have developed an experimental setup to discretise the motility of the plasmodium, and the motility was forced to be a stepwise one transition \cite{RefWorks:68}. In this way the behaviour of the plasmodium was similar to that of a two-dimensional cellular automaton. They analysed the motility of only a single Physarum with no source of attraction (food source). They postulated several models (transition rules) of Physarum movement based on the statistical results of several experiment runs.

\subsection*{Physarum electrical activity experiment}
\citeA{RefWorks:122, RefWorks:79} designed a real biological experiment where they measured the electrical activity of Physarum in the presence of stimuli (one food source) \cite{RefWorks:122,RefWorks:79}. The goal of the experiment was to show how the Physarum changes patterns of its electrical activity when exposed to attractants and repellents, based on the fact that Physarum learn and adapt to periodic changes in its environment \cite{RefWorks:93, RefWorks:59}. \citeA{RefWorks:121} demonstrated that the protoplasmic tubes of the Physarum showed current versus voltage characteristics that is consistent with ideal memristor-systems \cite{RefWorks:121}. 
\citeA{RefWorks:69} \cite{RefWorks:69} designed a real experiment similar to that of \citeA{RefWorks:122} \cite{RefWorks:122}; where they presented a bio-inspired memristor-based circuit maze-solving approach. 

\subsection*{Physarum solving the two-armed bandit problem experiment}
\begin{figure*}
\includegraphics[width=\textwidth]{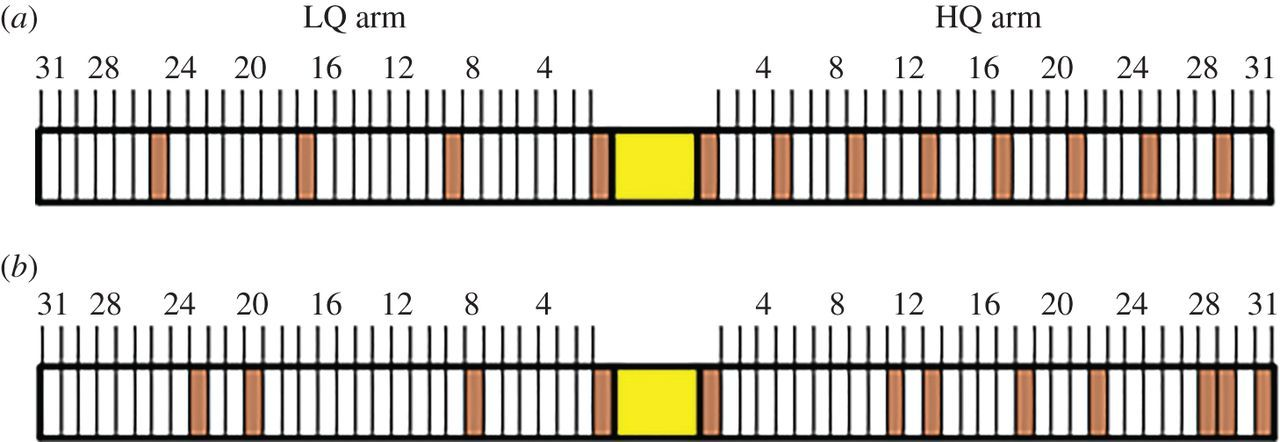}
\caption{Two-armed bandit experimental set-up for Physarum. Cell biomass was placed in the centre (yellow box). White boxes indicate blank agar sites (non-rewarding), brown boxes indicate oat-agar food sites (rewarding). Pictured here are the (a) 4e versus 8e treatment, where the LQ arm has evenly distributed reward sites, and the HQ arm has 8 evenly distributed reward sites, and (b) 4r versus 8r treatment, where the reward sites were distributed randomly \cite{RefWorks:124}.}
\label{fig:Literature_Review/Physarum_bandit_problem}
\end{figure*}

The two-armed bandit problem has previously only been used to study organisms with brains. Yet Physarum, a brainless unicellular organism, showed the ability of decision-making and solved the two-armed bandit problem. In this experiment, Physarum was challenged with a choice between two deferentially rewarding environments, where the arm with the greater number of food resources or higher quality was designated as the high-quality (HQ) arm, and the other arm with fewer food resources or low quality was designated as the low-quality (LQ) arm (Figure \ref{fig:Literature_Review/Physarum_bandit_problem}) \cite{RefWorks:124}. The outcome of this experiment was to demonstrate the Physarum decision-making abilities. Physarum always chose the high-quality arm, and it can make multi-objective foraging decisions. It compares the relative qualities of multiple options and combines the information on reward (frequency and magnitude) in order to make correct and adaptive decisions. This experiment provides insight into the fundamental principles of Physarum decision-making and information processing.


\subsection*{Physarum foraging behaviour in competition settings experiment}
\citeA{RefWorks:308} have designed a biological experiment to study the behaviour of Physarum under competition settings \cite{RefWorks:308}. The experiment intercalated two Physarum in a common environment (petri dish) where there was only one food resource available. The experimental results showed that the time taken by Physarum to find food depends on their hunger motivation. However, the time taken for a Physarum to start looking for food depended on its motivation and the motivation of its competitor. Physarum always initiates foraging behaviour quicker in the presence of competitors.

\begin{figure*}
\begin{center}
\includegraphics[scale=0.9]{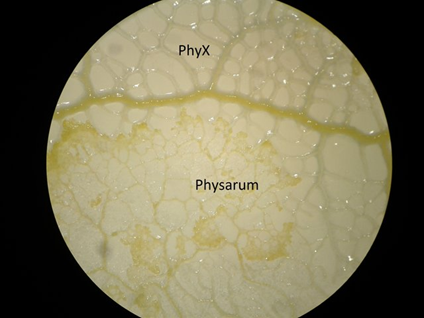}
\caption{Experiment with two agents: Physarum polycephalum could grow into branches of Badhamiautricularis \cite{RefWorks:316}.}
\label{fig:Model/Physarum_Competition_Schumann}
\end{center}
\end{figure*}

In another biological experiment by \citeA{RefWorks:316}, two strains were cultured in the same petri dish, the first was the usual Physarum Polycephalum plasmodium, and the second was another species called a Badhamiautricularis. Physarum Polycephalum definitely grows faster than Badhamiautricularis and overtakes more food resources, and could even grow into the branches of Badhamiautricularis, only if the Physarum inoculum was fatter (See Figure \ref{fig:Model/Physarum_Competition_Schumann}) \cite{RefWorks:316}.
Furthermore, if the invasive growth in front of Badhamiautricularis is well nourished by oat, it would easily overgrow the opposing tube system of Physarum Polycephalum. Thus, competitions between Physarum Polycephalum and Badhamiautricularis is based on Physarum power (type), mass, and the availability of nearby food resources.

\begin{figure*}[t!]
\includegraphics[width=\textwidth]{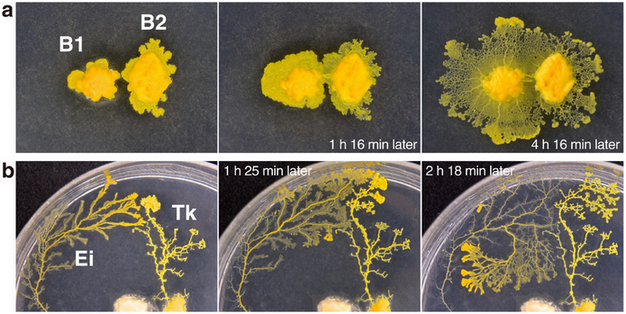}
\caption{Two typical encounter cases. (a) A fused case in which $B1$ and $B2$ completed fusion extremely smoothly to become a single individual. (b) An avoided case. As can be confirmed in this group of photographs, $E_i$ and $T_k$ encountered each other in at least five locations, recognised self and other, and chose to avoid the latter in all encounters. This avoidance behaviour was very clearly observed \cite{RefWorks:346}.}
\label{fig:Model/Allorecognition}
\end{figure*}

In a recent study by \citeA{RefWorks:346} \cite{RefWorks:346}, five geographical strains of Physarum with different genotypes were collected. In each experiment, two individual plasmodia on oat flakes were placed on $2\%$ agar in a round petri dish and were allowed to behave freely. Whether the individuals avoided or fused was recorded for all encounter cases. Allorecognition was defined as the time when the plasmodium came into contact with the other individual. Completion of allorecognition was defined as a change in behaviour (continuing straight, changing direction, or starting to fuse at the point of contact). The study has revealed that Physarum strictly identifies allogeneic individuals when encountering different individuals. The Allorecognition system in Physarum prioritises the avoidance and severely restricts fusion when encountering different individuals (Figure \ref{fig:Model/Allorecognition}) \cite{RefWorks:346}.

\section{Mathematical models for simulating Physarum foraging behaviour \label{sec:LiteratureReview/Mathematical_Models}}
Physarum biological experiments are extremely slow and time-consuming to be applied in real-world network design problems. It is rather better to use the meta-heuristic algorithms inspired by Physarum intelligent behaviour (as conducted in real biological experiments) to construct mathematical models. The existing models are simulating the intelligent behaviour of single Physarum, and have overlooked foraging behaviour of multiple Physarum under competitive settings. For this reason, we have presented our new model to simulate Physarum in competition, where multiple Physarum interact with each other and with their environments. We have summarised some of these existing mathematical models in Table \ref{tab:Literature_Review/Physarum_mathematical_model}.

\begin{table}[htbp]
  \centering
  \caption{Physarum mathematical models.}
  \label{tab:Literature_Review/Physarum_mathematical_model}
    \begin{tabular}{|l|l|p{10.715em}|p{13.215em}|}
    \toprule
    \rowcolor[rgb]{ .31,  .506,  .741} \multicolumn{1}{|c|}{\textbf{\# }} & \textbf{Author } & \textbf{Model } & \textbf{Application } \\
    \midrule
    \rowcolor[rgb]{ .816,  .847,  .91} 1     & \citeA{RefWorks:70}  & Hagen–Poiseuille Law and Kirchhoff Law  & Solving maze,  complex transport network.  \\
    \midrule
    \rowcolor[rgb]{ .914,  .929,  .957} 2     & \citeA{RefWorks:73}  & Reaction–Diffusion of Belousov–Zhabotinsky & Solve maze, graph problems and design logical gates.  \\
    \midrule
    \rowcolor[rgb]{ .816,  .847,  .91} 3     & \citeA{RefWorks:76}  & Cellular automaton  & Solve maze, Stainer minimum tree and spanning tree problems, and  transport network.  \\
    \midrule
    \rowcolor[rgb]{ .914,  .929,  .957} 4     & \citeA{RefWorks:78}  & Vacant particle based model  & Approximation of network formation.  \\
    \midrule
    \rowcolor[rgb]{ .816,  .847,  .91} 5     & \citeA{RefWorks:84}  & Multi-agent system  & Solve maze and optimize meta-heuristic algorithms.  \\
    \midrule
    \rowcolor[rgb]{ .914,  .929,  .957} 6     & \citeA{RefWorks:120}  & Memristor circuit  & Solve maze and transport networks.  \\
    \midrule
    \rowcolor[rgb]{ .816,  .847,  .91} 7     & \citeA{RefWorks:103}  & Cellular Automaton and the Reaction–Diffusion systems  & Solve maze and transport networks.  \\
    \midrule
    \rowcolor[rgb]{ .914,  .929,  .957} 8     & \citeA{RefWorks:368}  & Hexagonal Cellular Automaton and the Reaction–Diffusion systems  & Solve Mobile Wireless Sensor Networks and discrete multi-objective optimisation problems.  \\
    \bottomrule
    \end{tabular}%
\end{table}%


\subsection*{The flow-conductivity model}
The flow-conductivity model is based on Hagen-Poiseuille Law and Kirchhoff Law to describe the adaptive feature of path finding and the feedback between flux and conductivity of the protoplasm tubes \cite{RefWorks:70,RefWorks:321,RefWorks:71}. Experiments on Physarum led by \citeA{RefWorks:321} have proposed the mechanism of protoplasmic flow through Physarum's tubular veins, which is believed to account for Physarum's intelligence \cite{RefWorks:321}. The flow-conductivity model was first proposed by \citeA{RefWorks:70} and \citeA{RefWorks:71} to simulate Physarum foraging behaviour \cite{RefWorks:70,RefWorks:71}. This model can solve the shortest path-finding and the maze-solving process of Physarum. The model illustrates the feedback between the flux and the thick of protoplasmic tubes; first, open-ended tubes, which are not connected between the two food sources, are likely to disappear. Second, when two or more tubes connect the same two food sources, the longer tube is likely to disappear. The model was applied in dynamic navigation to design the railway network around Tokyo \cite{RefWorks:67}.

In this model, two terminals are representing Physarum (source/node), and the other terminal is food resource (sink/node). The protoplasm flows in every edge from the source node to the sink node. There is a pressure at each vertex, and the quantity of flux in each edge is proportional to the pressure difference between the two ends of these edges. Specifically, the flux $Q_{ij}$ in edge $(i,j)$ is given by the Hagen-Poiseuille equation below.
\begin{equation}
Q_{ij}=\frac{D_{ij}}{c_{ij}}(p_i-p_j)
\end{equation}
\begin{equation}
D_{ij}=\frac{\pi r^4_{ij}}{8\xi}
\end{equation}
where $D_{ij}$ is the edge conductivity, $c_{ij}$ is the edge length, $p_i$ and $p_j$ are pressures at vertices $i$ and $j$, $r_{ij}$ is the edge
radius, and $\xi$ is the viscosity coefficient.

\subsection*{Reaction-diffusion model}
\citeA{RefWorks:74} regards the Physarum as an encapsulated reaction-diffusion computer, and utilises a two-variable Oregonator equation to simulate the Physarum spanning tree construction \cite{RefWorks:73,RefWorks:74}. In this model, the wavefront is used to simulate the motion of Physarum, whose trajectory is steered by the gradient of chemo-attractants. It was treated as a bio-realised unconventional computer called "Physarum Machine" to solve maze problems, graph problems and design logical gates \cite{RefWorks:33}.

\subsection*{The cellular model \label{sec:LiteratureReview/Mathematical_Models/Cellular_Model}}
The cellular model was proposed by \citeA{RefWorks:76} \cite{RefWorks:76}. Given a planar lattice, and every lattice site has various states: the inside (state 1) is surrounded by a boundary (state 2) in a lattice outside (state 0). In the foraging phase, there is cell invasion of the outside with softening of the membrane of Physarum. The protoplasmic flow toward the softened area, which leads to a re-organisation of the distribution of the cytoskeleton.
This model was applied to simulate the amoebic motion and solve the classical Steiner tree problem in planes \cite{RefWorks:76,RefWorks:77}. Moreover, other researchers have developed a cellular automata model based on reaction diffusion to simulate the behaviour of Physarum \cite{RefWorks:85,RefWorks:103,RefWorks:323}.

\subsection*{The multi-agent model}
\citeA{RefWorks:324} has been proposed a multi-agent, where Physarum is thought to consist of a population of particle-like agents \cite{RefWorks:324}. Each agent senses and deposits trails as it moves towards the nearby stimulus within a 2D diffusive lattice. In this model, the structure of the Physarum network is indicated by the collective pattern of the positions of agents, and the protoplasmic flow is represented by the collective movement of agents. Furthermore, \citeA{RefWorks:325} improved the initial multi-agent model by adding a memory module to each agent \cite{RefWorks:325}. This improved model is more flexible and adaptive, and it approximates the behaviours of Physarum more closely.

\citeA{RefWorks:84} proposed a self-organised system modelling approach in which two types of agents are used for simulating both the search (exploration) and the contraction (exploitation) of Physarum in foraging behaviours \cite{RefWorks:84}. In this model, the body comprises a synthesis module and a motion module, and each sensor is armed with a trail sampling module and a chemo-nutrient sampling module.

\subsection*{Physarum Competition Model}
\citeA{RefWorks:368} proposed a novel model to imitate the complex patterns observed in Physarum polycephalum generated in competition settings \cite{RefWorks:368}. This new model is based on hexagonal Cellular Automata (CA) and Reaction-Diffusion (RD) systems. This is the first time Physarum has been simulated in a 2-D hexagonal grid that is more applicable to Physarum natural diffusion in a circular pattern to equidistant cells. All other models considered either Von-Neumann (4 adjacent neighbours) \cite{RefWorks:68} or Moore neighbourhoods (8 adjacent neighbours) \cite{RefWorks:103}. However, in Von-Neumann model, diagonal diffusion of Physarum can still occur, while in Moore model the neighbourhoods are not equidistant. In this competition model multiple Physarum interact with each other and with their environment, each Physarum has its autonomous behaviour: it compares information on reward determined by food resources’mass and quality, negative effects of competing neighbours according to their mass, and hunger motivation in order to make correct and adaptive decisions. They believe that competition among different Physarum individuals can lead to the emergence of a complex global behaviour, far beyond the capability of individual Physarum. The individual skills of competition are more efficient to achieve an optimal balance between exploration and exploitation and maintain population diversity.

\section{Physarum-Inspired Applications \label{sec:LiteratureReview/Physarum_Applications}}
In this section, we will address the most important question "What Physarum can offer to computing?". Many Physarum-inspired algorithms have been developed and proved to have great potential to solve various optimisation problems using simple heuristics. In this context we will not be restricted to graph optimisation problems as previous reviews \cite{RefWorks:213,RefWorks:214,RefWorks:33,RefWorks:387}, we will open the horizon and through light to more recent applications. We will address this issue by briefly reviewing some of the existing researches on these Physarum-inspired applications.


\subsection{Physarum-Inspired Algorithms for Graph-Optimisation Problems}
Physarum protoplasmic flux is changing continuously with the change of environment in its foraging process. This characteristic allows Physarum to have great potentials in dealing with graph-optimisation problems which are considered the main application.
Physarum network design has attracted the attention of many researchers as it demonstrated excellent performance in network construction without central consciousness during the process of foraging. The Physarum solver is based on positive feedback where the tubes that are more suitable for transporting the nutrients will grow bigger and will be of less resistance, while the tubes which do not transport enough nutrients will vanish and disappear. This feedback mechanism helps to maximise the number of nutrient sources and to minimise transportation costs \cite{RefWorks:66, RefWorks:345}. The Physarum solver constructs networks by making some nodes in the network "sources" and cytoplasmic streaming to others “sinks”. So there is a great difference between the way that Physarum solves the shortest path problem and the traditional methods, including the Dijkstra algorithm \cite{RefWorks:327}.


Many mathematical models were proposed to simulate the intelligent behaviour of Physarum (as discussed in Section \ref{sec:LiteratureReview/Mathematical_Models}). The algorithms based on these models were able to find the shortest path in directed and undirected networks. \citeA{RefWorks:34} were the first to
show how this simple organism has the ability to find the shortest path between two points in a labyrinth \cite{RefWorks:34}.
Subsequent research has confirmed and broadened the range of its computation abilities to spatial representations of various graph problems \cite{RefWorks:213}. It showed that the Physarum's network geometry met the requirements of a smart network: short tubes, close connections among all the branches, and tolerance to dynamic changes. \citeA{RefWorks:67} designed a Physarum bio-inspired networks similar to the Tokyo rail system \cite{RefWorks:67}. The resulting networks are both efficient and robust.

A lot of Physarum-inspired algorithms (PAs) have been proposed to solve challenging network optimisation problems, such as the travelling salesman problem \cite{RefWorks:95} and the Steiner tree problems \cite{RefWorks:295,RefWorks:330}, transport network design and simulation \cite{RefWorks:36, RefWorks:102}, spanning tree approximation \cite{RefWorks:75}, and vehicle routing problems \cite{RefWorks:328}. Recent examples of the Physarum application include: designing supply chain networks \cite{RefWorks:92}, community detection \cite{RefWorks:98}, and discrete multi-objective optimisation problems \cite{RefWorks:307}. For a detailed discussion on the existing methods and applications refer to \cite{RefWorks:213,RefWorks:214}.

These popular Physarum-inspired Algorithms (PAs) have proven its potential in solving challenging network optimisation problems \cite{RefWorks:45,RefWorks:213}. However, some network optimisation problems remain unsolved. New techniques are required to address the large scale of the next-generation networks, where centralised control of communication becomes impractical. Physarum distributed intelligence may inform the design of an adaptive, robust and spatial infrastructure networks with decentralised control systems \cite{RefWorks:214}. We have proposed a Physarum competition model \cite{RefWorks:368}, where multiple Physarum with autonomous behaviours react to each other and with their environment without central control to achieve efficient exploration of the whole system evolving to an optimal global network, this has allowed us to deal with the increasingly proposed networks scenarios with multiple sources and multiple sinks. In our previous work, we have presented a Physarum-inspired competition algorithms for mobile wireless sensor networks, where multiple Physarum (as represented by sensors) will sense the surrounding environment, and compete over multiple food resources (as represented by interest points). These algorithms have demonstrated their promising performance in solving node deployment \cite{RefWorks:220} and connectivity restoration even in harsh environment \cite{RefWorks:219}.

These network graph-optimisation problems are typically based on the following four strategies:
\begin{itemize}
 \item 
 One source node and one sink node: It was first proposed by \citeA{RefWorks:34} after performing his famous experiments showing that Physarum was able to find the shortest route through a maze \cite{RefWorks:34}. \citeA{RefWorks:329} solved the travelling salesman problem \cite{RefWorks:329}. \citeA{RefWorks:213} accelerated its optimisation process by intentionally removing the edges with a stable decreasing flow \cite{RefWorks:213}.
 
 \item
 Multiple source nodes and one sink node: this strategy is to select one terminal to be the sink node and then select the other terminals to be source nodes. It has been applied by \citeA{RefWorks:295} to solve the classical Steiner tree problem in graphs \cite{RefWorks:295}. It has also been used to solve the prize-collecting Steiner tree problem and the node-weighted Steiner tree problem \cite{RefWorks:331}.
 \item
 One source node and multiple sink nodes: this strategy is to select one terminal to be the source node and then select the other terminals to be sink nodes. It was first used by \citeA{RefWorks:332} to design transportation networks with fluctuating traffic distributions \cite{RefWorks:332}.
 \item
 Multiple source nodes and multiple sink nodes: this strategy is to select multiple terminals to be the source nodes and multiple terminals to be the sink nodes. It was recently proposed by \citeA{RefWorks:213} to solve the supply chain network design problem \cite{RefWorks:213}.
\end{itemize}

\subsection{Evolutionary Algorithm Optimisation (Hybrid Models)}
Prior knowledge plays a vital role in the computational efficiency of evolutionary algorithms (e.g., Genetic Algorithm, and Ant Colony). Taking advantage of Physarum powerful computational capabilities, such as morphological diversity \cite{RefWorks:76} and positive feedback loop \cite{RefWorks:260}, these characteristics have been used to optimise some evolutionary algorithms to improve its efficiency and robustness \cite{RefWorks:333, RefWorks:98}.


Ant colony optimisation (ACO) algorithms have been shown to provide an approximate solution for NP-hard problems existing in many real-world applications. However, premature convergence has significantly reduced the performance of these algorithms. \citeA{RefWorks:333} proposed an optimisation strategy for updating the pheromone matrix in ant colony algorithms based on a Physarum mathematical model \cite{RefWorks:333}. This strategy has accelerated the positive feedback process in ACO, for solving NP-hard problems such as travelling salesman problem (TSP) and 0/1 knapsack problem, which contributed to the quick convergence of the optimal solution \cite{RefWorks:94}. Later on \citeA{RefWorks:98} has incorporated Physarum-inspired initialisation to optimise the genetic algorithm, ant colony optimisation algorithm and Markov clustering algorithm for solving community detection problems \cite{RefWorks:98}.


\subsection{Biological Computing and Physarum Logic Gates}
Boolean logic which describes binary arithmetic is fundamental to computer science as electronic logic gates form the basis of digital operations in computers. Organism based Bio-Logic gates have been attempted using cell constituent (bacteria) as transducers \cite{RefWorks:337}. Bacteria have many drawbacks, mainly due to the fragility, short life, limited temperature, and pH conditions. Also, bacteria will often not grow on specific substrates which would be ideal for the cell-transducer interface. Yeast and wild fungi are offering the advantage of high growth rate and the ability to grow on a broad range of surface substrates used for cell-transducer interface \cite{RefWorks:338}. Moreover, yeast can survive for over a long time after dehydration and could be re-hydrated when required.

Like other fungi and yeast, Physarum is accessible to culture on moist filter paper or agar and resist dehydration for a long time. This is why it can be considered as a prospective experimental prototype of biological computers which does not require sophisticated support. In standard electric devices, we deal with electrical signals to code information. However, in a Physarum biological device instead of electrical signals, the calculation process is performed by using the Physarum chemotaxis to food \cite{RefWorks:33, RefWorks:339}. 


Physarum as a method of biological computing has been extensively studied in the PhyChip project that ran between 2013 and 2016 "Physarum chip: growing computers from slime mould" \cite{RefWorks:326}. A Physarum chip is formed of a living network of protoplasmic tubes that acts as an active non-linear transducer of information, while templates of tubes coated with conductor act as fast information channels. The symbolic-logical, mathematical and programming aspects of the Physarum chip have been studied by \citeA{RefWorks:316} \cite{RefWorks:316}. Physarum was also used as a Boolean gate, where the presence and absence of Physarum in a given locus of space is equivalent to logic values 1 and 0, respectively \cite{RefWorks:39}. The Physarum chip is expected to solve a wide range of computation tasks, including graph optimisation, logic and arithmetical computing \cite{RefWorks:341}.

The EU-funded PhySense project "Physarum Sensor: Biosensor for Citizen Scientists" is an extension of the PhyChip project. This project showed that Physarum is an ideal biological substrate that could be used as a biosensor that converts a biological response into an electrical signal, providing a unique fusion of living and digital technology. The PhySense software calculates any changes in the frequency and amplitude of oscillations in the tubular structures of Physarum. The aim of this project is developing marketable low-cost biosensors for various applications, including environmental monitoring and health \cite{RefWorks:355} \footnote{More information: PhySense project website: \url{www.physense.eu/}}.

\section{Conclusion \label{sec:LiteratureReview/Conclusion}}
By studying Physarum foraging behaviour and translating that behaviour into mathematical models, we increase our understanding of how to inspire from biology to develop Physarum bio-inspired algorithms can solve many challenging real world problems. Physarum polycephalum is an example of plasmodial slime moulds. The primitive intelligence of Physarum polycephalum is mostly demonstrated during its plasmodium stage (a large multi-nucleated single cell).  that consists of a single cell amoeba-like organism. Physarum senses gradients of chemo-attractants and repellents and forms a yellowish vascular network in search of nutrition. A stimulus triggers the release of a signalling molecule cyclic adenosine monophosphate (cAMP) which starts cytoplasmic streaming. This generates a feedback loop; the higher the rate of cytoplasmic streaming is, the thicker the vein becomes. The Physarum foraging behaviour consists of two simultaneous self-organised processes of expansion (exploration) and shrinkage (exploitation). Just like social insects and animals, Physarum too exhibits swarm intelligence; it shares many features of collective behaviour such as synchronisation, communication, positive feedback, leadership, and response thresholds. There is increasing evidence that a simple organism like Physarum has complex social behaviours including cooperation and competition. Physarum is capable of making complex foraging decisions based on trade-offs between risks, hunger level and food patch quality. The skills of individual competitors are effective methods for inspiration to develop intelligent systems and to provide solutions for decision-making problems.

Physarum may not have brains, but the advantages of Physarum unconventional computational capabilities, as morphological diversity and positive feedback loop, have great potentials for solving many NP-hard problems. Physarum, as a simple organism, has the ability to find the minimum-length between two points in solving the maze problem and discover the shortest path in real-world networks such as the Tokyo railway network using simple heuristics. Much research has confirmed and broadened the range of its computation abilities to spatial representations of various graph optimisation problems. Physarum-inspired initialisation of other bio-inspired techniques has the ability to accelerate convergence and improve the searching capability of evolutionary algorithms (e.g., Genetic Algorithm, and Ant Colony) in terms of accuracy and computational cost. Physarum as a biological model has been studied in the PhyChip and PhySense projects to develop marketable and low-cost biosensors for various applications including environmental monitoring and health.Physarum can be considered one of the biological models of unconventional computation capable of making a programmable Physarum machine.

\section{Acknowledgement}
Dr. Abubakr Awad research was supported by Elphinstone PhD Scholarship, University of Aberdeen. Dr. Wei Pang, Prof. David Lusseau, and Prof. George M. Coghill were supported by the Royal Society International Exchange program (Grant Ref IE160806).

\bibliography{main.bib}  

\end{document}